\documentclass[lettersize,journal]{IEEEtran}
\usepackage{amsmath,amsfonts}
\usepackage{algorithmic}
\usepackage{algorithm}
\usepackage{array}
\usepackage[caption=false,font=footnotesize,labelfont=rm,textfont=rm]{subfig}
\usepackage{textcomp}
\usepackage{stfloats}
\usepackage{url}
\usepackage{verbatim}
\usepackage{graphicx}
\usepackage{cite}
\usepackage{multirow}
\usepackage{multicol}
\usepackage[table]{xcolor} 
\usepackage{colortbl}
\usepackage{makecell}
\usepackage{booktabs}
\usepackage{threeparttable}
\usepackage[colorlinks,
            linkcolor=red,
            anchorcolor=blue,
            citecolor=green
            ]{hyperref}
\begin{document}

\title{Fuzzy-aware Loss for Source-free Domain Adaptation in Visual Emotion Recognition}

% \author{Anonymous submission}

\author{Ying Zheng$^\dagger$, Yiyi Zhang$^\dagger$, Yi Wang,~\IEEEmembership{Member,~IEEE}, and Lap-Pui Chau$^\ast$,~\IEEEmembership{Fellow,~IEEE}% <-this % stops a space
\thanks{The research work was partly conducted in the JC STEM Lab of Machine Learning and Computer Vision funded by The Hong Kong Jockey Club Charities Trust. This work was supported in part by the National Natural Science Foundation of China (No. 62106236).}% <-this % stops a space
\thanks{Ying Zheng, Yi Wang and Lap-Pui Chau are with the Department of Electrical and Electronic Engineering, The Hong Kong Polytechnic University, Hong Kong, China. E-mail: \{ying1.zheng, yi-eie.wang, lap-pui.chau\}@polyu.edu.hk.}% <-this % stops a space
\thanks{Yiyi Zhang is with the Department of Computer Science and Engineering, The Chinese University of Hong Kong, Hong Kong, China. E-mail: yyzhang24@cse.cuhk.edu.hk.}% <-this % stops a space
\thanks{$\dagger$ denotes equal contribution; $\ast$ denotes corresponding author.}}

\markboth{Fuzzy-aware Loss for Source-free Domain Adaptation in Visual Emotion Recognition}%
{Shell \MakeLowercase{\textit{et al.}}: A Sample Article Using IEEEtran.cls for IEEE Journals}

\IEEEpubid{0000--0000/00\$00.00~\copyright~2025 IEEE. Personal use of this material is permitted.}

\maketitle

\begin{abstract}
Source-free domain adaptation in visual emotion recognition (SFDA-VER) is a highly challenging task that requires adapting VER models to the target domain without relying on source data, which is of great significance for data privacy protection. However, due to the unignorable disparities between visual emotion data and traditional image classification data, existing SFDA methods perform poorly on this task. In this paper, we investigate the SFDA-VER task from a fuzzy perspective and identify two key issues: fuzzy emotion labels and fuzzy pseudo-labels. These issues arise from the inherent uncertainty of emotion annotations and the potential mispredictions in pseudo-labels. To address these issues, we propose a novel fuzzy-aware loss (FAL) to enable the VER model to better learn and adapt to new domains under fuzzy labels. Specifically, FAL modifies the standard cross entropy loss and focuses on adjusting the losses of non-predicted categories, which prevents a large number of uncertain or incorrect predictions from overwhelming the VER model during adaptation. In addition, we provide a theoretical analysis of FAL and prove its robustness in handling the noise in generated pseudo-labels. Extensive experiments on 26 domain adaptation sub-tasks across three benchmark datasets demonstrate the effectiveness of our method.
% Code is available at: \url{https://github.com/zhengyinghit/FAL}.
\end{abstract}


\section{Introduction}
\label{sec:introduction}
\IEEEPARstart{V}{isual} emotion recognition (VER) has emerged as a crucial research direction in computer vision and affective computing  \cite{zhao2021affective,shu2024fuzzy,han2024fmfn}. With the explosive growth of social media and digital content, VER has shown significant potential for applications such as human-robot interaction \cite{chen2020fuzzy} \cite{zheng2024survey}, social media analytics \cite{machajdik2010affective}, and enhancing entertainment experiences \cite{zheng2021sketch}. By interpreting the emotional cues embedded in visual data like images and videos, VER enables machines to more effectively comprehend and react to human emotions, ultimately enhancing user experiences. However, despite significant progress in recent years, VER still encounters several issues in real-world scenarios. One of the main issues is the diversity of visual data distribution. Visual emotional data is usually collected from various sources and environments, leading to substantial differences in data distribution. For example, users from different social platforms may come from diverse cultural backgrounds and use different shooting conditions, resulting in distinct emotional expression styles and features. This distribution difference makes it difficult for models trained on one dataset to perform well on new datasets. Furthermore, visual emotional data often raises privacy concerns, and the difficulty and cost associated with acquiring large-scale annotated data further limit the generalization ability of VER models \cite{lin2020multi}.

To address the aforementioned issues, source-free domain adaptation (SFDA) presents a promising solution. SFDA seeks to facilitate a trained model to learn and adapt to the data distribution of an unlabeled target domain without relying on source domain data \cite{li2024comprehensive}. This approach allows the model to achieve strong performance in the target domain while alleviating privacy concerns related to the use of source domain data. By leveraging SFDA, models can effectively perform emotion recognition in the target domain, thereby enhancing the practicality and reliability of VER systems. However, existing SFDA methods are primarily designed for general image classification tasks and do not perform well when directly applied to VER. This is mainly due to the specific complexity of VER. Firstly, the diversity and subtlety of emotional expressions make the emotion recognition task more complex than general classification tasks. Secondly, the subjectivity of emotional data increases the difficulty of model adaptation. Existing SFDA methods usually do not consider these specific emotional characteristics, leading to poor performance in the SFDA-VER task.

\begin{table}[t]
\centering
\caption{The percentage (\%) of certain and uncertain samples in manually annotated visual emotion dataset. A certain sample means that several annotators tag the same emotional label to this image, while an uncertain sample indicates the existence of disagreement among the annotators.}
\begin{tabular}{lcccc}
\toprule
\multirow{2}{*}{Dataset} & \multicolumn{2}{c}{Positive} & \multicolumn{2}{c}{Negative}\\
{} & {Certain} & {Uncertain} & {Certain} & {Uncertain}\\
\midrule
Twitter I (T1) \cite{you2015robust} & 75.6 & 24.4 & 60.2 & 39.8 \\
Instagram (In) \cite{katsurai2016image} & 49.5 & 50.5 & 43.1 & 56.9 \\
Flickr (Fl) \cite{katsurai2016image} & 44.8 & 55.2 & 31.6 & 68.4 \\
\bottomrule
\end{tabular}
\label{tab:uncertain}
\end{table}

\IEEEpubidadjcol

To tackle the challenges of SFDA-VER, we first delve into this task from a fuzzy perspective. Specifically, through the analysis of visual emotion data, we identify two aspects of fuzzy problems in SFDA-VER. (1) The fuzzy emotion label problem, refers to the inherent uncertainty in emotion labels. The annotation of emotional data is usually subjective, so existing VER datasets generally determine the emotional category labels of images through voting, which results in a certain degree of ambiguity in the labels themselves. Table \ref{tab:uncertain} shows the percentage of certain and uncertain samples across three datasets that contain two emotional categories, i.e., positive and negative. Uncertainty in emotional labels is prevalent across these datasets, particularly in the negative category of the Flickr \cite{katsurai2016image} dataset, where 68.4\% of the samples have inconsistent annotations. (2) The fuzzy pseudo-label problem, which means the inaccuracy of the model's prediction. In existing SFDA work, many approaches rely on pseudo-labels generated by the source-pretrained model for adaptation. The pseudo-labels may be incorrect, which also contributes to fuzzy problems. We alternate between using EmoSet and FI as the source and target domain, and test the classification performance of the source model on the target dataset, as shown in Table \ref{tab:accuracy}. The accuracy of each category reflects the reliability of the pseudo-labels generated by the source model. Overall, the reliability of pseudo-labels is not very high, especially the accuracy of the fear category is only 26.2\% when F$\rightarrow$E. Potential incorrect pseudo-labels can be viewed as noise, which may lead to overfitting on noisy data during adaptation.

\begin{table}[t]
\centering
\caption{The source-pretrained model's classification accuracy (\%) for each category in the target domain dataset. E and F represent EmoSet \cite{yang2023emoset} and FI \cite{you2016building}, respectively, and ``$\rightarrow$'' indicates that the model is trained on the source domain (left) and tested on the target domain (right). To better format the table, we shorten the names of each emotional category.}
\begin{tabular}{ccccccccc}
\toprule
{Task} & {Amu.} & {Ang.} & {Awe} & {Con.} & {Dis.} & {Exc.} & {Fea.} & {Sad.}\\
\midrule
E$\rightarrow$F & 67.9 & 40.0 & 66.3 & 44.2 & 50.8 & 48.8 & 61.7 & 53.9 \\
F$\rightarrow$E & 39.0 & 27.2 & 61.1 & 69.6 & 76.4 & 65.8 & 26.2 & 36.1 \\
\bottomrule
\end{tabular}
\label{tab:accuracy}
\end{table}

Based on the above observations, we propose a fuzzy-aware learning method to address the fuzzy problems, in which the core is a novel fuzzy-aware loss (FAL) function. FAL combines the standard cross entropy loss with a fuzzy-aware loss term. This new loss term adaptively calibrates the losses from other categories to give attention to the fuzzy problems, thereby achieving more robust domain adaptation. We also provide a theoretical analysis of FAL and demonstrate its robustness to noise in generated pseudo-labels. Furthermore, we establish the connection between FAL and two classical loss functions, i.e., focal loss \cite{lin2017focal} and reverse cross entropy loss \cite{wang2019symmetric}, to enhance the understanding of FAL.

To validate the effectiveness of the proposed method, we conduct extensive experiments comparing eight SFDA methods and eight loss functions across multiple public datasets. These datasets include 2-class sentiment datasets, 8-class emotion datasets, and the Office-Home object recognition dataset, encompassing a total of 26 different domain adaptation sub-tasks. Our experimental results reveal that existing SFDA methods exhibit poor performance on the SFDA-VER task, while loss functions perform better. However, these loss functions struggle on general SFDA datasets. This indicates a disparity between SFDA-VER and traditional SFDA tasks, which leads to varying performance for SFDA methods and loss functions. In contrast, our method not only achieves the best performance on the SFDA-VER task but also demonstrates comparable results to state-of-the-art methods on the challenging Office-Home dataset.

In summary, our contributions are as follows:
\begin{itemize}
    \item We explore the SFDA-VER problem from a fuzzy perspective, which holds significant implications for data privacy protection.
    \item We propose a novel loss function, termed fuzzy-aware loss (FAL), which effectively addresses the fuzzy problems in SFDA-VER. Additionally, we provide a theoretical analysis of the robustness of FAL.
    \item We conduct extensive experiments to demonstrate the superior performance of FAL in the SFDA-VER task, and we also validate the effectiveness of FAL on the general SFDA dataset.
\end{itemize}

\section{Related Work}
\label{sec:relatedwork}
In this section, we will introduce the related literature in the field of visual emotion recognition and source-free domain adaptation. Given that the proposed method is closely related to loss functions designed for learning with noisy labels, we will also outline the related research in this area.

\subsection{Visual Emotion Recognition}
Existing VER work can be classified into two main categories based on the study subjects: human-centered \cite{kosti2020context} \cite{yang2024robust} and human-agnostic \cite{yang2018visual} \cite{yang2023emoset}. Human-centered emotion recognition (HCER) focuses on identifying the emotional states of the target person, necessitating the presence of human subjects within the visual content. Traditional HCER methods predominantly utilize human-related attributes, such as facial expressions \cite{li2020deep}, body postures \cite{bhattacharya2020step}, and acoustic behaviors \cite{mittal2020m3er}, to predict people's emotional state. Recently, advancements in psychological research have spurred interest in context-aware emotion recognition (CARE) \cite{kosti2017emotion} \cite{yang2023context}. CARE aims to enhance emotional features by extracting rich contextual information from images. For example, Yang \textit{et al.} \cite{yang2024robust} introduce a counterfactual emotion inference (CLEF) framework to mitigate context bias. CLEF employs a causal graph and a context branch to eliminate the direct effects of contextual bias, thereby enhancing the robustness of visual emotion recognition.

Human-agnostic emotion recognition (HAER) investigates the emotional responses elicited by images, focusing on the emotions conveyed by the visual content without imposing restrictions on the image's subject matter. HAER does not require the presence of humans in images, thus encompassing a broader range of content and serving as a generalized form of HCER. To determine an image's emotion, it is essential to establish the connection between visual cues and abstract emotions. Early deep learning methods \cite{xu2014visual} \cite{you2015robust} leverage convolutional neural networks (CNNs) to extract image features, followed by emotion classification using fully connected classifiers. For instance, You \textit{et al.} \cite{you2015robust} introduce a progressive CNN (PCNN) approach that incrementally fine-tunes a CNN model with a dataset containing generated pseudo-labels. While these methods connect the holistic image features to emotions directly, they often overlook local image cues. To address this limitation, You \textit{et al.} \cite{you2017visual} and Yang \textit{et al.} \cite{yang2018visual} propose identifying significant emotional regions and constructing classification models based on these regions. To further integrate global and local emotional information, Yang \textit{et al.} \cite{yang2021solver} develop a method using a graph convolutional network (GCN) to model the relationships between scenes and objects, thereby inferring the image's emotion through these associations. Xu \textit{et al.} \cite{xu2022mdan} design a multi-level dependent attention network (MDAN), which connects image features with emotional semantics at multiple levels, enhancing emotion recognition outcomes. Recently, Zhao \textit{et al.} \cite{zhao2024to} propose a novel emotion recognition evaluation method based on the Mikel emotion wheel from psychology, which considers the distances between different emotions on the wheel to capture emotional nuances more accurately.

Furthermore, several approaches incorporate supplementary information to enhance emotion recognition in images. Borth \textit{et al.} \cite{borth2013sentibank} develop a visual concept detector library named SentiBank, which significantly improves emotion recognition performance by utilizing adjective-noun pairs. Yang \textit{et al.} \cite{yang2023emoset} demonstrate on the large-scale sentiment recognition dataset EmoSet that models integrating emotion attribute prediction can achieve superior emotion recognition results. Recently, with the remarkable performance of large-scale pre-trained models such as CLIP \cite{radford2021learning}, there has been increased interest in leveraging large language models to provide additional guidance for emotion recognition \cite{deng2023simple} \cite{pan2024multi}. For instance, Deng \textit{et al.} \cite{deng2023simple} put forward a prompt-based fine-tuning method that combines visual and language features to capture richer emotional cues. However, there remains an absence of SFDA research within the emotion recognition domain. Given the importance of data security and privacy protection, we believe that the exploration of SFDA-VER will facilitate the application of VER technology in the future.

\subsection{Source-free Domain Adaptation}
The difference between SFDA and standard unsupervised domain adaptation (UDA) lies in their data utilization strategies. UDA methods \cite{zhang2022well} \cite{shi2024unsupervised} leverage both source and target data during model training, whereas SFDA operates without access to source data, relying exclusively on a pre-trained source model and target data for updates. This characteristic of SFDA underscores its importance in maintaining data privacy. Since the inception of SFDA by Liang \textit{et al.} \cite{liang2020we}, the field has attracted considerable interest within the research community, leading to notable advancements.

In the pioneering SHOT \cite{liang2020we} framework, the authors freeze the classifier module of the source model and then update the feature extractor through an information maximization loss and self-supervised pseudo-labeling. SHOT exhibits excellent performance in the SFDA task for object recognition and has been adopted and extended by lots of subsequent works \cite{liang2021source,yang2021generalized,yang2021exploiting,yang2022attracting,lee2022confidence,zhang2022divide,karim2023c,xia2024discriminative}. Building upon SHOT, Liang \textit{et al.} \cite{liang2021source} propose the SHOT++ method, which further improve SHOT's accuracy through a label transfer strategy and semi-supervised learning. Yang \textit{et al.} \cite{yang2022attracting} conceptualize SFDA as an unsupervised clustering problem. They accomplish effective clustering of local neighborhood features within the feature space by optimizing an objective focused on prediction consistency. Zhang \textit{et al.} \cite{zhang2022divide} introduce a paradigm termed Divide and Contrast (DaC), which divides target data into source-like and target-specific samples. Within an adaptive contrastive learning framework, DaC fulfills distinct global and local learning objectives for each sample group. Xia \textit{et al.} \cite{xia2024discriminative} propose a discriminative pattern calibration (DPC) method, which leverages the discriminative patterns of target images to enhance feature representation for SFDA.

Most of the aforementioned SFDA works rely on generated pseudo-labels and new loss functions to guide the learning of specific features or local structures in the target data. The proposed FAL method also aligns with this learning paradigm. However, despite the extensive body of work on SFDA, its applicability to the field of VER remains unexplored. Our experimental results show that most of the existing SFDA methods do not perform well when transferred to the VER task. Therefore, we posit the critical need for methods specifically tailored to the distinctive attributes of VER.

\subsection{Loss Functions for Learning with Noisy Labels}
During the process of model adaptation to the target domain, it is typically necessary to utilize pseudo-labels generated on target images. However, these labels inevitably contain many errors, making learning based on partially incorrect labels close to the problem of learning with noisy labels (LNL). Within the field of LNL, a critical research direction involves designing loss functions that are robust to noisy labels.

Previous research has theoretically proven that the widely used cross entropy (CE) loss function lacks robustness in the presence of noisy labels \cite{ghosh2017robust}. Consequently, subsequent studies \cite{zhang2018generalized,wang2019symmetric,liu2020peer,ma2020normalized,ye2023active,zhou2023asymmetric} have focused on developing modified versions of the CE loss to facilitate more robust learning. These methods usually mitigate the impact of noisy labels on model training by incorporating additional objective functions or calibrating the loss values. For example, the generalized cross entropy (GCE) \cite{zhang2018generalized} loss integrates the mean absolute error (MAE) with CE, balancing the two terms through a hyperparameter. Wang \textit{et al.} \cite{wang2019symmetric} propose a symmetric cross entropy (SCE) loss that combines CE with a robust reverse cross entropy (RCE) loss. Considering the challenge of identifying a convex loss function under symmetric conditions, Zhou \textit{et al.} \cite{zhou2023asymmetric} present a new family of loss functions, namely asymmetric loss functions, which preserve some beneficial properties of convex loss functions to facilitate subsequent network optimization.

Existing loss functions for LNL have improved the robustness of models in noisy environments. However, since the target data addressed by these methods differ significantly from VER data in terms of emotional semantics and noise distribution, they cannot be directly applied to solve the problem of SFDA-VER. This discrepancy has been verified in the experiments of Section \ref{sec:experiments}. Hence, how to design a robust loss function suitable for SFDA-VER remains an urgent challenge for further investigation.

\section{Method}
\label{sec:method}
In this section, we begin by introducing some preliminary knowledge related to SFDA and conduct a thorough analysis of the drawbacks of CE in SFDA-VER. Then, we provide a detailed description of the proposed FAL method, accompanied by theoretical analysis and discussion.

\subsection{Preliminaries}
In this paper, we focus on the classical closed-set setting of SFDA, where the source and target domains share the same $K$ classes. In the SFDA task, we first define two datasets: the source domain dataset $\mathcal{D}_s$ containing $N_s$ samples, and the target domain dataset $\mathcal{D}_t$ containing $N_t$ samples. Specifically, the source domain dataset is represented as $\mathcal{D}_s = \{(\mathbf{x}_i^s, y_i^s)\}_{i=1}^{N_s}$, where $\mathbf{x}_i^s$ denotes the samples from the source domain and $y_i^s$ indicates the corresponding class labels. The target domain dataset is denoted as $\mathcal{D}_t = \{\mathbf{x}_i^t\}_{i=1}^{N_t}$, containing only samples $\mathbf{x}_i^t$ from the target domain, without providing the true class labels. Given a model $\mathcal{M}_s$ that has been pre-trained on the source domain dataset $\mathcal{D}_s$ using supervised learning, the objective is to adapt this model to maximize its recognition performance on the target domain $\mathcal{D}_t$ without utilizing source data. This adaptation results in an adapted model $M_t$ such that $\mathcal{M}_t(\mathbf{x}_i^t)$ can accurately predict the class labels of the target domain samples.

In general, the model $\mathcal{M}_t$ is composed of two components: a feature extractor $\mathcal{F}_t$ and a classifier $\mathcal{C}_t$. The feature extractor $\mathcal{F}_t$ processes an input sample $\mathbf{x}^t$ from the target domain, yielding a feature vector $z = \mathcal{F}_t(\mathbf{x}^t)$ that resides within a feature space of dimension $h$. Following this, the classifier $\mathcal{C}_t$ takes the feature vector $z$ as input and produces an output $p = \delta(\mathcal{C}_t(z)) \in \mathbb{R}^K$, where $\delta$ denotes the softmax function. The output $p$ satisfies the condition $\sum_{i=1}^{K} p(i|\mathbf{x}^t)=1$ and each $p(i|\mathbf{x}^t)$ represents the predicted probability that the sample $\mathbf{x}^t$ belongs to class $i$. Subsequently, the pseudo-label $\hat{y}$ for $\mathbf{x}^t$ is obtained by identifying the class with the highest probability: 
\begin{equation}
    \hat{y} = \arg\max_i \, p(i|\mathbf{x}^t).
    \label{eq:pseudolabel}
\end{equation}
Regarding the imprecise pseudo-labels generated on target data, our goal is to formulate a robust loss function for the task of SFDA-VER. This new loss aims to mitigate the negative impact of ambiguous labels on model updates, thereby improving the model's stability and performance in handling uncertain or ambiguous inputs.

\subsection{Revisit Cross Entropy Loss}
We denote the label distribution for sample $\mathbf{x}^t$ by $q(i|\mathbf{x}^t)$ and $\sum_{i=1}^{K}q{\{i|\mathbf{x}^t\}}=1$. In the commonly used one-hot encoding for labels, we have $q(\hat{y}|\mathbf{x}^t)=1$ and $q(i|\mathbf{x}^t)=0$ for all $i \neq \hat{y}$. For notational convenience, we use $p_i$ and $q_i$	in place of $p(i|\mathbf{x}^t)$ and $q(i|\mathbf{x}^t)$ in all the following equations. The cross entropy (CE) loss is formulated as:
\begin{equation}
    \ell_{ce} = -\sum_{i=1}^{K}q_{i}\log (p_i),
    \label{eq:celoss1}
\end{equation}
Recall that when $i=\hat{y}$, $q_{i}=1$; otherwise $q_{i}=0$. Therefore, Eq. (\ref{eq:celoss1}) can be rewritten as: 
\begin{equation}
    \ell_{ce} = -\log (p_{\hat{y}}) - \sum_{i \neq \hat{y}}q_{i}\log (p_i),
    \label{eq:celoss2}
\end{equation}
where the second term equals to $0$. From Eq. (\ref{eq:celoss2}), it is evident that the CE loss only focuses on the difference between the model's predictions and the pseudo-labels. The objective is to make the model's predictions gradually approach the pseudo-labels. This method is effective on datasets with clean labels, as it helps guide the model in optimizing towards the true labels. However, in the SFDA-VER task, due to the fuzzy problems, where the generated pseudo-labels are not entirely accurate, the CE loss may lead the model to overfit on incorrect pseudo-labels. As a result, the CE loss is not an ideal option in this context.

\subsection{Fuzzy-aware Learning}
In SFDA-VER, there are two types of fuzzy problems: fuzzy emotion labels and fuzzy pseudo-labels. Both issues can result in the pseudo-labels that the model uses during domain adaptation being potentially inaccurate, i.e., $\hat{y}$ may not be equal to the true label $y$. As we have observed in the analysis of CE loss, CE is ineffective in dealing with the fuzzy problems because it solely concentrates on the loss associated with pseudo-labels, neglecting the importance of true labels that may exist in other classes. This limitation results in poor performance when handling noisy data, as CE cannot differentiate between true and pseudo-labels. Consequently, there is a need for a more robust loss function that accounts for the impact of both true labels and pseudo-labels, thereby improving the model's performance in the SFDA-VER task.

To incorporate the potential impact of true labels belonging to other classes into the loss function, an alternative solution is to introduce a non-zero coefficient $\lambda_{i}$ to replace $q_i$ in Eq. (\ref{eq:celoss2}). Since $\lambda_{i}$ is not zero, the outputs of other classes will also contribute to the model's updating. This approach can increase the model's attention to samples that may be misclassified during adaptation, thereby improving its robustness. Then, we have a new variant of the CE loss:
\begin{equation}
    \ell_{ce}^\ast = -\log (p_{\hat{y}}) - \sum_{i \neq \hat{y}}\lambda_{i}\log (p_i),
    \label{eq:celoss3}
\end{equation}
where $\lambda_{i}$ is still an undetermined coefficient. In the Shannon entropy of information theory, $p_i\log(p_i)$ represents the information content of each possible value, reflecting the contribution of each event's probability to the overall uncertainty. Inspired by this, we set $\lambda_{i}=p_{i}$, then we have:
\begin{align}
    \ell_{fal} &= -\log (p_{\hat{y}}) - \sum_{i \neq \hat{y}}p_i\log (p_i) \nonumber \\
    &= - \sum_{i=1}^{K}q_{i}\log (p_i) - \sum_{i=1}^{K}p_i\log (p_i) + p_{\hat{y}}\log (p_{\hat{y}}) \nonumber \\
    &= - \sum_{i=1}^{K}q_{i}\log (p_i) - \sum_{i=1}^{K}p_i\log (p_i) + \sum_{i=1}^{K} p_i q_i \log (p_i) \nonumber \\
    &= - \sum_{i=1}^{K} \lbrack (1-p_i)q_i + p_i \rbrack \log(p_i)
    \label{eq:faloss1}
\end{align}
In the above equation, introducing $\lambda_{i}=p_{i}$ enables the new loss $\ell_{fal}$ better to handle data uncertainty with pseudo-labels. This allows the model to take into account the fuzzy problems between different classes during decision-making. Therefore, we refer to $\ell_{fal}$ as fuzzy-aware loss (FAL) and the domain adaptation learning based on $\ell_{fal}$ as fuzzy-aware learning.

Additionally, we introduce a weighting coefficient to adjust the loss for each category during the adaptation process. As a commonly used method to alleviate class imbalance \cite{zheng2020deep}, weighted loss allows for fine-tuning the impact of each category on the overall loss, enhancing the model's ability to adapt to varying category distributions. Specifically, we maintain a memory bank $B$ to store the predictions from the last testing round for all samples. Then, the weight for class $i$ is defined as,
\begin{equation}
    w_i = \frac{M}{n_i},
\end{equation}
in which
\begin{equation}
    M = \frac{1}{K} \sum_{i=1}^{K} n_i,
\end{equation}
where $n_i$ represents the number of samples within the memory bank $B$ that belong to the $i$-th category. Thus, we obtain a weight-balanced variant of FAL:
\begin{equation}
    \ell_{fal} = - \sum_{i=1}^{K} w_i \lbrack (1-p_i)q_i + p_i \rbrack \log(p_i)
    \label{eq:faloss2}
\end{equation}

\subsection{Theoretical Analysis}
As illustrated in the first equality of Eq. (\ref{eq:faloss1}), the proposed FAL consists of two parts: the traditional CE loss and a new fuzzy-aware loss term $\ell_{ft}=- \sum_{i \neq \hat{y}}p_i\log (p_i)$. As highlighted in references \cite{ghosh2017robust} \cite{zhang2018generalized}, the CE loss is unbounded and lacks robustness against noisy data. Therefore, the robustness of FAL primarily stems from the fuzzy-aware term. Next, we will theoretically explore the boundedness and robustness of the fuzzy-aware term.

\textbf{Boundedness analysis}. Before analyzing the boundedness of $\ell_{ft}$, we first consider the function $g(p) = -p \log(p) $ for $ p \in (0, 1)$. When $ p \to 0^+ $, although $ \log(p) \to -\infty $, but with the product of $p$, $g(p) \to 0 $. When $ p \to 1^- $, $ \log(p) \to 0 $, hence $g(p) \to 0 $. Next, we compute the derivative of $g(p)$:
\begin{equation}
    g'(p) = -(\log(p) + 1)
    \label{eq:derivative}
\end{equation}
and set $ g'(p) = 0 $ to find critical points:
\begin{equation}
    \log(p) = -1 \quad \Rightarrow \quad p = e^{-1}.
    \label{eq:extreme}
\end{equation}
Then, we evaluate $g(p)$ at the critical point $p = e^{-1}$:
\begin{equation}
    g\left(e^{-1}\right) = -e^{-1} \log\left(e^{-1}\right) = e^{-1}.
    \label{eq:critical}
\end{equation}
Thus, the maximum and minimum values of $g(p)$ on $(0, 1)$ are $e^{-1}$ and 0, respectively.

As $\ell_{ft}$ equals to $\sum_{i \neq \hat{y}} g(p_i)$, we have:
\begin{equation}
    0 < \ell_{ft} \leq (K-1)e^{-1}.
    \label{eq:bounded}
\end{equation}
This demonstrates the boundedness of $\ell_{ft}$, which makes it more robust compared to unbounded losses such as CE.

\textbf{Robustness analysis}. We consider the pseudo-label $\hat{y}$ of a sample $\mathbf{x}$ as a noisy label, with the true label represented by $y$. Given any classifier $f$ and loss function $\ell_{ft}$, we define $R(f)=\mathbb{E}_{\mathbf{x}, y}\ell_{ft}$ as the risk of the classifier $f$ with clean labels and $R^\eta(f)=\mathbb{E}_{\mathbf{x}, \hat{y}}\ell_{ft}$ as the risk under noisy label rate $\eta$. Let $\tilde{f}$ and $f^\ast$ be the global minimizers of $R^\eta(f)$ and $R(f)$, respectively. Following \cite{wang2019symmetric} \cite{ghosh2017robust}, a loss function is deemed noise-tolerant if it ensures that the probability of misclassification at the optimal decision boundary is identical for clean and noisy data. In the following, we will introduce an assumption for the SFDA-VER task and prove $\ell_{ft}$ is robust under label noises.

\hypertarget{target1}{\underline{\textit{Assumption 1}}}. \textit{Let $\eta_{yi}$ be the probability that a sample $\mathbf{x}$ with a true label $y$ is predicted as class $i$ by the source model. The target domain dataset $\mathcal{D}_t = \{\mathbf{x}_i^t, \hat{y}_i\}_{i=1}^{N_t}$ in SFDA-VER is clean-labels-dominant, i.e., it satisfies that $\forall i \neq y$, $\eta_{yi} < 1 - \eta_y$ with $\sum_{i \neq y}\eta_{yi}=\eta_y$.}

In SFDA-VER, the source model exhibits significant performance differences across different categories, resulting in uneven noise distribution among various categories in the target domain dataset $\mathcal{D}_t = \{\mathbf{x}_i^t, \hat{y}_i\}_{i=1}^{N_t}$. This can be approximately regarded as asymmetric or class-dependent noise. Here, $1 - \eta_y$ represents the probability of the label being correct, while $\eta_{yi} < 1 - \eta_y$ indicates that for category $y$, the probability of correctly predicting the label is greater than the probability of being incorrectly predicted as another class. As discussed in \cite{zhou2023asymmetric}, a method capable of learning the correct classifier when clean labels are not dominant will fail in scenarios where clean labels are dominant. This occurs because the learned classifier will likely predict samples as belonging to the less dominant class instead of the true dominant class. Therefore, to train a reliable classifier, the training dataset needs to satisfy the assumption of clean-labels-domination.
\begin{figure}[htbp]
    \centering
    {\includegraphics[width=0.9\linewidth]{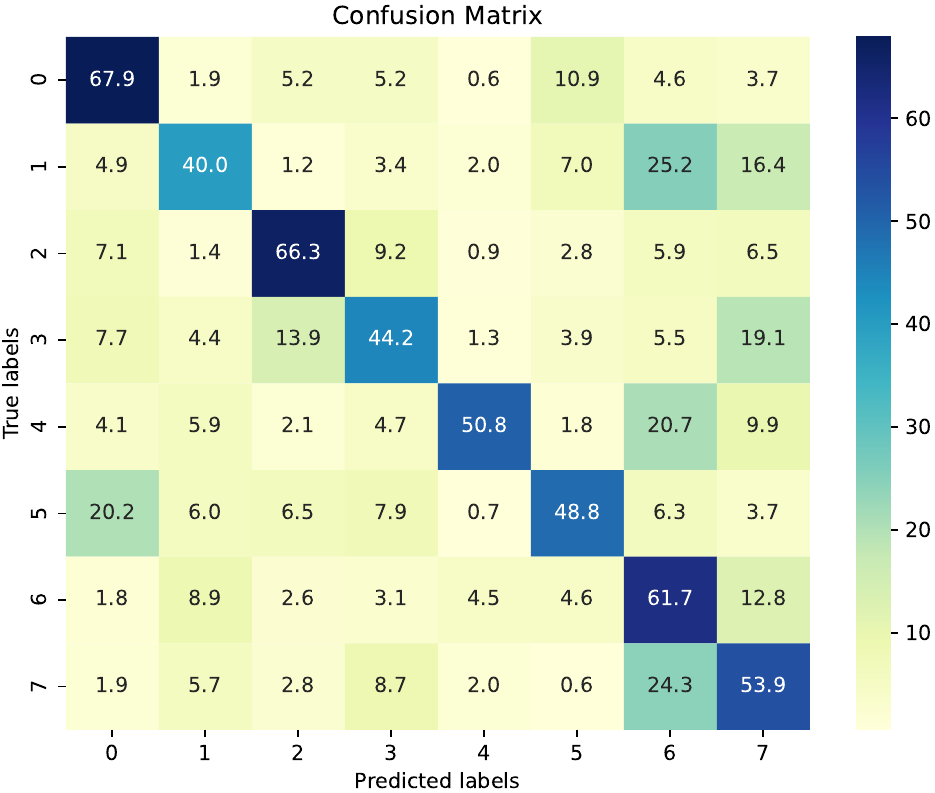}}
    \caption{The confusion matrix (\%) for directly testing the source model (ResNet-50) on the target data domain (EmoSet$\rightarrow$FI).}
    \label{fig:emoset2fi}
\end{figure}

An empirical study of the clean-labels-domination assumption is illustrated in Fig. \ref{fig:emoset2fi}. We utilized the source model (ResNet-50) trained on EmoSet to directly test its performance on FI, thereby obtaining the confusion matrix for eight emotion categories. From this figure, it can be observed that the confusion matrix is diagonal-dominant, which means that it satisfies condition $\eta_{yi} < 1 - \eta_y$. It should be noted that Assumption \hyperlink{target1}{1} is relatively strict and may not hold for each category in every task due to variations of source models and target data domains. However, based on practical experience from SFDA-VER, this assumption is generally valid in most cases. Therefore, noise-tolerant loss functions remain effective.

\hypertarget{target2}{\underline{\textit{Theorem 1}}}. \textit{When $\eta_{yi} < 1 - \eta_y$ with $\sum_{i \neq y}\eta_{yi}=\eta_y$, then $0 \leq R^{\eta}(f^\ast) - R^{\eta}(\tilde{f}) \leq C_K$, where $C_K=K(K - 1) e^{-1} \mathbb{E}_{\mathbf{x}, y} (1 - \eta_y)$.}

\textit{Proof.} Following the proof of Theorem 1 in \cite{wang2019symmetric} and the proof of Theorem 2.3 in \cite{wei2023mitigating}, for asymmetric or class-dependent label noise, we have:
\begin{equation}
\begin{split}
R^\eta(f) &= \mathbb{E}_{\mathbf{x}, \hat{y}} \ell_{ft}(f(\mathbf{x}), \hat{y}) = \mathbb{E}_{\mathbf{x}} \mathbb{E}_{\hat{y}|\mathbf{x}} \mathbb{E}_{\hat{y}|\mathbf{x}, y} \ell_{ft}(f(\mathbf{x}), \hat{y}) \\
&= \mathbb{E}_{\mathbf{x}} \mathbb{E}_{y|\mathbf{x}} \Big[ (1 - \eta_y) \ell_{ft}(f(\mathbf{x}), y) + \sum_{i \neq y} \eta_{yi} \ell_{ft}(f(\mathbf{x}), i) \Big] \\
&= \mathbb{E}_{\mathbf{x}, y} \Big[ (1 - \eta_y) \Big( \sum_{i=1}^K \ell_{ft}(f(\mathbf{x}), i) - \sum_{i \neq y} \ell_{ft}(f(\mathbf{x}), i) \Big) \Big] \\&\quad+ \mathbb{E}_{\mathbf{x}, y} \Big[ \sum_{i \neq y} \eta_{yi} \ell_{ft}(f(\mathbf{x}), i) \Big].
\end{split}
\label{eq:proof1}
\end{equation}
Recall that $0 < \ell_{ft} \leq (K-1)e^{-1}$ in Eq. (\ref{eq:bounded}), we have:
\begin{equation}
\begin{split}
R^\eta(f) &\leq \mathbb{E}_{\mathbf{x}, y} \Big[ (1 - \eta_y) \Big( K (K - 1) e^{-1} - \sum_{i \neq y} \ell_{ft}(f(\mathbf{x}), i) \Big) \Big] \\&\quad+ \mathbb{E}_{\mathbf{x}, y} \Big[ \sum_{i \neq y} \eta_{yi} \ell_{ft}(f(\mathbf{x}), i) \Big] \\
&= C_K - \mathbb{E}_{\mathbf{x}, y} \Big[ \sum_{i \neq y} \varphi_i \ell_{ft}(f(\mathbf{x}), i) \Big],
\end{split}
\label{eq:proof2}
\end{equation}
where $C_K=K(K - 1) e^{-1} \mathbb{E}_{\mathbf{x}, y} (1 - \eta_y)$ and $\varphi_i=(1-\eta_y-\eta_{yi})$. On the other hand, we have:
\begin{equation}
\begin{split}
R^\eta(f) > - \mathbb{E}_{\mathbf{x}, y} \Big[ \sum_{i \neq y} \varphi_i \ell_{ft}(f(\mathbf{x}), i) \Big].
\end{split}
\label{eq:proof3}
\end{equation}
Hence,
\begin{align}
&R^{\eta}(f^\ast) - R^{\eta}(\tilde{f}) \nonumber \leq \\ C_K + \mathbb{E}_{\mathbf{x}, y} \Big[ \sum_{i \neq y} \varphi_i &\Big(\ell_{ft}(\tilde{f}(\mathbf{x}), i) - \ell_{ft}(f^\ast(\mathbf{x}), i)\Big) \Big].
\label{eq:proof4}
\end{align}
Due to the assumption that $\eta_{yi} < 1 - \eta_y$, it follows that $\varphi_i=(1-\eta_y-\eta_{yi}) > 0$. We represent the lower bound of the loss $\ell_{ft}$ with $b^-$ and the upper bound with $b^+$. As $f^\ast$ is the global minimizer of $R(f)$, we assume $R(f^\ast) = b^-$. In this instance, $\ell_{ft}(f^*(\mathbf{x}), i)$ should be equal to the upper bound $b^+$ for all $i \neq y$. Thus, we have:
\begin{equation}
\begin{split}
\mathbb{E}_{\mathbf{x}, y} \Big[ \sum_{i \neq y} \varphi_i \Big(\ell_{ft}(\tilde{f}(\mathbf{x}), i) - \ell_{ft}(f^*(\mathbf{x}), i)\Big) \Big] \leq 0.
\end{split}
\label{eq:proof5}
\end{equation}
As $\tilde{f}$ is the global minimizer of $R^{\eta}(f)$, $R^{\eta}(f^\ast) - R^{\eta}(\tilde{f}) \geq 0$. Finally, from Eq. (\ref{eq:proof4}), we have:
\begin{equation}
\begin{split}
0 \leq R^{\eta}(f^\ast) - R^{\eta}(\tilde{f}) \leq C_K.
\end{split}
\label{eq:proof6}
\end{equation}
This completes the proof.

According to Theorem \hyperlink{target2}{1}, we can conclude that the risk difference between the global minimizers $\tilde{f}$ and $f^\ast$ under noisy and clean labels consistently falls within a specific boundary. Therefore, the minimizer of the true risk can be approximately regarded as the minimizer of the risk under noisy data. This demonstrates that $\ell_{ft}$ is noise-tolerant under asymmetric or class-dependent label noise.

% \textbf{Gradient analysis}.

% \begin{figure}[htbp]
%     \centering
%     {\includegraphics[width=1.0\linewidth]{figures/-plogp.png}}
%     \caption{-plogp.}
%     \label{fig:-plogp}
% \end{figure}

\subsection{Discussion}
In this section, we will connect two well-known loss functions—focal loss and reverse cross entropy loss—to FAL. This discussion can enhance the understanding of our method. For the clarity of discussion, we reformulate Eq. (\ref{eq:faloss1}) into a two-term expression:
\begin{equation}
    \ell_{fal} = - \sum_{i=1}^{K} (1-p_i)q_i \log(p_i) - \sum_{i=1}^{K} p_i \log(p_i)
    \label{eq:faloss3}
\end{equation}

\textbf{Focal loss} (FL), introduced by Lin \textit{et al.} \cite{lin2017focal} for improving the performance of dense object detection tasks, also serves as a powerful loss function for classification problems. FL is defined as:
\begin{equation}
    \ell_{fl} = - \sum_{i=1}^{K} (1 - p_i)^\gamma q_i \log(p_i),
\end{equation}
where $(1 - p_i)^\gamma$ is a modulating factor and $\gamma$ is a tunable focusing parameter. When $\gamma=0$, FL simplifies to CE, but as $\gamma$ increases, the effect of the modulating factor is enhanced, putting more emphasis on misclassified examples. When $\gamma=1$, FL becomes exactly the first term of FAL.

\textbf{Reverse cross entropy} (RCE) loss \cite{wang2019symmetric} is proposed to be combined with CE to create the symmetric cross entropy (SCE) loss, which has been demonstrated to be robust against label noise. The RCE loss is defined as:
\begin{equation}
    \ell_{rce} = -\sum_{i=1}^{K} p_i \log(q_i)
\end{equation}
Due to the label $q_i$ being one-hot encoded, there might be an issue with values of zero inside the logarithm. To address this problem, the authors set $\log (0)$ to a negative value $A$, e.g., $A=-4$. For notational convenience, we define $A^\ast$:
\begin{equation}
    A^\ast = \log(q_i) = \left\{
\begin{array}{ll}
0 & \text{if } i = \hat{y} \\
A & \text{if } i \neq \hat{y},
\end{array}
\right.
\end{equation}
and rewrite $\ell_{rce}=-\sum_{i=1}^{K} p_i A^\ast$. In the second term of Eq. (\ref{eq:faloss3}), $\log(p_i)$ is a small value that changes with $p_i$. If we treat it as the constant $A^\ast$, this term becomes equivalent to RCE.

Thus, we have elucidated the relationship between FAL and both FL and RCE, and found that FAL can be regarded as the combination of learning components associated with these two loss functions. FAL incorporates the advantages of FL and RCE and does not contain manually set hyperparameters, which can achieve better performance on the SFDA-VER task.

\section{Experiments} % 3.5 pages
\label{sec:experiments}
In this section, we experimentally validate the proposed method and compare it with several contemporary state-of-the-art techniques. We begin by outlining the datasets, the methods used for comparison, and the implementation details. Next, we present the results of visual emotion recognition, as well as the experimental results for object recognition. Finally, we provide a comprehensive analysis and discussion of our method.

\subsection{Datasets}
To explore the effectiveness of SFDA-VER methods, we adopt six widely used visual emotion recognition datasets, as shown in Table \ref{tab:dataset}. These datasets' emotion categories are based on two psychological models: the sentiment model and the Mikels model \cite{mikels2005emotional}. The sentiment model classifies emotions into two categories: positive and negative. The Mikels model categorizes emotions into eight groups: amusement, awe, contentment, excitement, anger, disgust, fear, and sadness. In constructing the evaluation benchmarks, we removed a small number of problematic images to more accurately evaluate the performance of the methods, including duplicate and damaged images. Below is a summary of these datasets: 

\begin{itemize}
    \item \textit{2-class Sentiment datasets}: (1) \underline{Twitter I} \cite{you2015robust}, abbreviated as T1, consists of 1,251 images. (2) \underline{Twitter II} \cite{borth2013large}, abbreviated as T2, consists of 545 images. (3) \underline{Instagram} \cite{katsurai2016image}, abbreviated as In, consists of 42,848 images. (4) \underline{Flickr} \cite{katsurai2016image}, abbreviated as Fl, consists of 60,729 images. All images in these datasets are collected from social websites using keyword searches aligned with emotion categories. In these four datasets, any dataset can be chosen as the source domain and another as the target domain, resulting in a total of 12 domain adaptation tasks.
\end{itemize}

\begin{itemize}
    \item \textit{8-class Emotion datasets}: (1) \underline{FI} \cite{you2016building} is constructed from Flickr and Instagram and contains 21,829 images. (2) \underline{EmoSet} \cite{yang2023emoset} is the largest visual emotion recognition dataset available to date, comprising a total of 118,102 images. We alternate the two datasets as source and target domains, thus forming two domain adaptation tasks. Although the number of sub-tasks is less than the Sentiment datasets, this benchmark is more challenging due to the inclusion of more emotion categories. This can be seen from the difference in accuracy observed in the experimental results.
\end{itemize}

\begin{table}[htbp]
\centering
\caption{Statistics of visual emotion recognition datasets.}
\resizebox{1.0\linewidth}{!}{
\begin{tabular}{lrcl}
\toprule
{Dataset} & {\#Image} & {\makecell{Model\\(\#category)}} & {Categories}\\
\midrule
Twitter I (T1) \cite{you2015robust} & 1,251 & \multirow{4}{*}{\makecell{Sentiment\\(2)}} & \multirow{4}{*}{positive, negative}\\
Twitter II (T2) \cite{borth2013large} & 545 & {} & {}\\
Instagram (In) \cite{katsurai2016image} & 42,848 & {} & {}\\
Flickr (Fl) \cite{katsurai2016image} & 60,729 & {} & {}\\
\midrule
FI \cite{you2016building} & 21,829 & \multirow{2}{*}{\makecell{Mikels\\(8)}} & \multirow{2}{*}{\makecell[l]{amusement, awe, contentment, excitement,\\anger, disgust, fear, sadness}}\\
EmoSet \cite{yang2023emoset} & 118,102 & {} & {}\\
\bottomrule
\end{tabular}
}
\label{tab:dataset}
\end{table}

To further verify the generality of the proposed method, we perform experiments on the Office-Home dataset \cite{venkateswara2017deep} to assess our FAL. Office-Home is a widely used and highly challenging object recognition dataset within the SFDA field. It comprises four image domains: Artistic images (Ar), Clipart (Cl), Product images (Pr), and Real-world images (Rw), encompassing a total of 12 domain adaptation tasks. This dataset includes 65 categories of everyday objects, amounting to 15,500 images in total.

\subsection{Comparison Methods}
We select and implement several representative methods for comparison in our experiments. These methods consist of the most advanced SFDA techniques and a range of up-to-date loss functions. The following is a summary of these comparison methods and the academic conferences or journals where they were published:

\begin{itemize}
    \item \textit{SFDA methods}: (1) \underline{SHOT} (ICML 2020) \cite{liang2020we}, which learns domain-specific features with the information maximization loss. (2) \underline{SHOT++} (TPAMI 2021) \cite{liang2021source}, which upgrades SHOT through a labeling transfer strategy. (3) \underline{G-SFDA} (ICCV 2021) \cite{yang2021generalized}, which groups target features that are semantically similar through local structure clustering. (4) \underline{NRC} (NeurIPS 2021) \cite{yang2021exploiting}, which encourages semantic consistency among samples with high neighborhood affinity through neighborhood reciprocity clustering. (5) \underline{AaD} (NeurIPS 2022) \cite{yang2022attracting}, which utilizes a two-term objective function that encourages local neighbor predictions to be consistent while pushing predictions for dissimilar features as far apart as possible. (6) \underline{DaC} (NeurIPS 2022) \cite{zhang2022divide}, which divides the target data into two subsets: source-like and target-specific, and then achieves class-wise domain adaptation through adaptive contrastive learning. (7) \underline{CoWA-JMDS} (ICML 2022) \cite{lee2022confidence}, which distinguishes the importance of samples in the target domain based on the joint model-data structure (JMDS) score. (8) \underline{C-SFDA} (CVPR 2023) \cite{karim2023c}, which prevents noise propagation in pseudo-labels through curriculum learning.
\end{itemize}

\begin{itemize}
    \item \textit{Loss functions}: (1) \underline{Focal loss} (ICCV 2017) \cite{lin2017focal}, which adds a modulating factor to the standard CE loss to address the class imbalance problem. (2) \underline{GCE} (NeurIPS 2018) \cite{zhang2018generalized}, which uses the negative Box-Cox transformation as a noise-robust loss function and can be seen as a generalization of mean absolute error (MAE) and categorical cross entropy (CCE). (3) \underline{NLNL} (ICCV 2019) \cite{kim2019nlnl}, which proposes negative learning to prevent deep models from overfitting to noisy data. (4) \underline{SCE} (ICCV 2019) \cite{wang2019symmetric}, which combines RCE with CE to form a robust symmetric loss. (5) \underline{APL} (ICML 2020) \cite{ma2020normalized}, which normalizes existing loss functions to make them robust to noise and combines active and passive losses to form more powerful losses. (6) \underline{PolyLoss} (ICLR 2022) \cite{leng2022polyloss}, which represents loss functions as a linear combination of polynomial functions and can adjust the importance of different polynomial bases. (7) \underline{ANL} (NeurIPS 2023) \cite{ye2023active}, which replaces the MAE in APL with normalized negative loss functions. (8) \underline{AUL} (TPAMI 2023) \cite{zhou2023asymmetric}, which introduces the asymmetry in loss functions to make them more robust to noise.
\end{itemize}

\subsection{Implementation Details}
To ensure a fair comparison with related work, we utilize a pre-trained ResNet-50 \cite{he2016deep} model on ImageNet as the backbone network. Following the literature \cite{liang2020we,liang2021source,yang2021generalized,yang2021exploiting,yang2022attracting}, we replace the original fully connected (FC) layer with a bottleneck layer and a new FC classifier layer. The bottleneck layer comprises a fully connected layer followed by a batch normalization (BN) layer. During network training, we employ the stochastic gradient descent (SGD) optimizer with a momentum of 0.9. The initial learning rate is set to 1e-3 for the backbone network and 1e-2 for both the bottleneck and FC layers, using a batch size of 64. All images are resized to $256\times256$ and randomly cropped to $224\times224$ for network input. In the testing phase, the center crop is employed to obtain $224\times224$ inputs. When training the source model on the Sentiment, Emotion, and Office-Home datasets, the maximum number of training epochs is empirically set to 10, 10, and 50, respectively. In the optimization of the target model, the maximum epochs are set to 5, 5, and 15, respectively.

\begin{table*}[t]
\centering
\caption{Classification accuracies (\%) on the 2-class Sentiment datasets (ResNet-50). We use \textbf{bold} to indicate the best results and \underline{underline} to highlight the second-best results.}
\resizebox{1.0\linewidth}{!}{
\begin{tabular}{llcccccccccccc>{\columncolor{gray!20}}c}
\toprule
{Method (Source→Target)} & Venue & {T1$\rightarrow$T2} & {T1$\rightarrow$In} & {T1$\rightarrow$Fl} & {T2$\rightarrow$T1} & {T2$\rightarrow$In} & {T2$\rightarrow$Fl} & {In$\rightarrow$T1} & {In$\rightarrow$T2} & {In$\rightarrow$Fl} & {Fl$\rightarrow$T1 } & {Fl$\rightarrow$T2} & {Fl$\rightarrow$In } & {Avg.}\\
\midrule
\textit{SFDA methods}\\
SHOT \cite{liang2020we} & ICML20 & 68.1 & 63.7 & 65.3 & 65.2 & 53.0 & 52.6 & 79.6 & 67.3 & 73.2 & 81.6 & 69.4 & 74.1 & 67.7\\
SHOT++ \cite{liang2021source} & TPAMI21 & 65.1 & 62.1 & 67.4 & 65.2 & 51.4 & 51.8 & 79.2 & 64.4 & 76.0 & 80.5 & 69.2 & 74.7 & 67.3\\
G-SFDA \cite{yang2021generalized} & ICCV21 & 64.8 & 77.2 & \textbf{79.3} & 61.1 & 44.8 & \textbf{79.3} & 79.5 & 65.0 & 64.9 & 74.6 & \textbf{69.7} & 77.2 & 69.8\\
NRC \cite{yang2021exploiting} & NeurIPS21 & 64.2 & 60.7 & 54.1 & 60.6 & 52.7 & 49.2 & \underline{80.3} & 66.8 & 67.3 & 79.3 & 68.4 & 63.5 & 63.9\\
AaD \cite{yang2022attracting} & NeurIPS22 & 63.9 & 59.9 & 57.7 & 61.0 & 59.5 & 53.6 & 77.6 & 66.2 & 56.1 & 79.1 & 66.2 & 62.4 & 63.6\\
DaC \cite{zhang2022divide} & NeurIPS22 & 60.9 & 60.1 & 63.6 & 64.8 & 52.1 & 55.2 & 77.1 & 67.5 & 75.2 & 77.0 & 68.8 & 70.9 & 66.1\\
CoWA-JMDS \cite{lee2022confidence} & ICML22 & \textbf{69.2} & 74.8 & 75.7 & \textbf{70.2} & 61.5 & 67.1 & \textbf{80.7} & 67.5 & 83.2 & \textbf{81.1} & \textbf{69.7} & 81.1 & 73.5\\
C-SFDA \cite{karim2023c} & CVPR23 & 64.8 & 71.6 & 73.0 & 67.8 & 58.0 & 66.7 & 75.4 & 67.5 & 83.3 & 72.7 & 68.1 & 79.9 & 70.7\\
\midrule
\textit{loss functions}\\
Focal loss \cite{lin2017focal} & ICCV17 & 68.4 & 75.2 & 77.8 & 63.7 & 61.3 & 67.7 & 78.7 & \underline{68.3} & \underline{84.4} & 78.1 & 69.0 & 82.3 & 72.9\\
GCE \cite{zhang2018generalized} & NeurIPS18 & 68.1 & 75.9 & 78.3 & 66.4 & \textbf{66.1} & 65.4 & 78.5 & \textbf{68.4} & \textbf{84.5} & 77.9 & \underline{69.4} & \textbf{82.6} & 73.5\\
NLNL \cite{kim2019nlnl} & ICCV19 & 68.3 & 75.4 & 77.9 & 63.9 & 60.7 & 66.3 & 79.7 & \underline{68.3} & 84.2 & 79.1 & 68.3 & 82.3 & 72.9\\
SCE \cite{wang2019symmetric} & ICCV19 & 67.3 & 76.5 & \underline{78.4} & 64.0 & 58.8 & 64.9 & 79.7 & 67.7 & 83.8 & 80.0 & 68.8 & 82.1 & 72.7\\
APL \cite{ma2020normalized} & ICML20 & 67.7 & 77.0 & \textbf{79.3} & 62.9 & 62.9 & \underline{75.0} & 79.5 & 67.7 & 83.5 & 79.8 & 68.4 & 80.9 & \underline{73.7}\\
PolyLoss \cite{leng2022polyloss} & ICLR22 & 68.1 & 75.6 & 78.2 & 63.6 & 61.0 & 65.4 & 79.5 & \textbf{68.4} & 84.0 & 79.4 & 67.7 & 82.2 & 72.8\\
ANL \cite{ye2023active} & NeurIPS23 & 68.1 & \underline{77.8} & 77.0 & 65.2 & \underline{64.5} & 67.4 & 79.5 & 67.3 & 84.2 & \underline{80.8} & 68.1 & 81.4 & 73.4\\
AUL \cite{zhou2023asymmetric} & TPAMI23 & 68.1 & 77.2 & \textbf{79.3} & 61.8 & 55.7 & 70.2 & 79.6 & 67.7 & 83.2 & 80.7 & 68.6 & 80.9 & 72.8\\
\midrule
Source model only & - & 66.2 & 73.0 & 73.4 & 61.2 & 59.7 & 63.3 & 79.6 & 66.4 & 81.9 & 78.7 & 68.3 & 80.3 & 71.0\\
\textbf{FAL (ours)} & - & \underline{68.8} & \textbf{78.2} & 77.8 & \underline{69.7} & 64.1 & 69.9 & \underline{80.3} & \textbf{68.4} & \textbf{84.5} & 80.3 & 69.2 & \underline{82.5} & \textbf{74.5}\\
\midrule
Target-supervised & - & 77.1 & 84.8 & 86.2 & 82.1 & 84.8 & 86.2 & 82.1 & 77.1 & 86.2 & 82.1 & 77.1 & 84.8 & 82.6\\
\bottomrule
\end{tabular}
}
\label{tab:sentiment}
\end{table*}

\begin{table*}[htbp]
\centering
\caption{Classification accuracies (\%) on the 8-class Emotion datasets (ResNet-50). We use \textbf{bold} to indicate the best results and \underline{underline} to highlight the second-best results.}
\resizebox{0.9\linewidth}{!}{
\begin{tabular}{llcccccccc>{\columncolor{gray!20}}c}
\toprule
{Method (EmoSet$\leftrightarrow$FI)} & Venue & {Amusement} & {Anger} & {Awe} & {Contentment} & {Disgust} & {Excitement} & {Fear} & {Sadness} & {Per-class}\\
\midrule
\textit{SFDA methods}\\
SHOT \cite{liang2020we} & ICML20 & 60.8$\vert$\textbf{43.7} & 39.5$\vert$34.7 & 70.1$\vert$71.1 & 45.0$\vert$54.5 & 69.6$\vert$\underline{84.2} & 55.4$\vert$63.4 & 56.6$\vert$47.8 & 53.1$\vert$\underline{53.5} & \underline{56.2}$\vert$\underline{56.6}\\
SHOT++ \cite{liang2021source} & TPAMI21 & 66.5$\vert$37.5 & 36.3$\vert$32.2 & 68.6$\vert$\textbf{72.1} & 44.7$\vert$49.1 & 68.0$\vert$82.0 & 56.5$\vert$65.7 & 43.7$\vert$\textbf{49.6} & \underline{59.6}$\vert$\textbf{54.9} & 55.5$\vert$55.4\\
G-SFDA \cite{yang2021generalized} & ICCV21 & 64.9$\vert$31.1 & 46.1$\vert$25.7 & 66.5$\vert$63.6 & 38.2$\vert$58.9 & 58.0$\vert$77.0 & \underline{57.7}$\vert$66.6 & 55.8$\vert$17.7 & 35.3$\vert$40.8 & 52.8$\vert$47.7\\
NRC \cite{yang2021exploiting} & NeurIPS21 & 46.5$\vert$33.3 & \textbf{55.2}$\vert$\textbf{49.5} & 61.5$\vert$42.7 & 36.6$\vert$34.0 & 65.6$\vert$77.7 & 43.2$\vert$35.6 & 52.6$\vert$41.7 & 40.5$\vert$35.4 & 50.2$\vert$43.7\\ 
AaD \cite{yang2022attracting} & NeurIPS22 & 45.5$\vert$14.6 & \underline{51.8}$\vert$36.4 & 60.3$\vert$11.5 & 33.7$\vert$13.8 & \textbf{71.7}$\vert$81.6 & 41.9$\vert$39.6 & 52.4$\vert$24.1 & 41.3$\vert$39.4 & 49.8$\vert$32.6\\
DaC \cite{zhang2022divide} & NeurIPS22 & 64.3$\vert$35.8 & 29.9$\vert$20.6 & \textbf{71.2}$\vert$\underline{71.3} & 45.3$\vert$54.3 & 66.8$\vert$74.9 & \textbf{63.4}$\vert$69.3 & 39.7$\vert$28.4 & 53.7$\vert$41.5 & 54.3$\vert$49.5\\
CoWA-JMDS \cite{lee2022confidence} & ICML22 & 65.6$\vert$\underline{42.9} & 32.9$\vert$\underline{36.6} & 66.9$\vert$65.4 & 42.9$\vert$43.8 & \underline{71.6}$\vert$79.0 & 54.7$\vert$\textbf{76.0} & 41.1$\vert$37.4 & 55.0$\vert$39.8 & 53.8$\vert$52.6\\  
C-SFDA \cite{karim2023c} & CVPR23 & \textbf{72.2}$\vert$41.6 & 34.9$\vert$30.8 & 70.1$\vert$54.4 & 44.2$\vert$72.2 & 61.2$\vert$71.1 & 54.7$\vert$61.0 & 35.2$\vert$25.4 & 39.2$\vert$32.3 & 51.5$\vert$48.6\\
\midrule
\textit{loss functions}\\
Focal loss \cite{lin2017focal} & ICCV17 & 67.5$\vert$40.4 & 40.6$\vert$27.3 & 66.9$\vert$61.8 & 45.8$\vert$72.2 & 50.0$\vert$78.4 & 49.6$\vert$66.7 & \underline{61.8}$\vert$28.4 & 54.1$\vert$36.8 & 54.5$\vert$51.5\\ 
GCE \cite{zhang2018generalized} & NeurIPS18 & \underline{69.8}$\vert$41.4 & 31.7$\vert$20.5 & 68.0$\vert$63.3 & 49.1$\vert$77.7 & 54.1$\vert$83.0 & 53.1$\vert$69.4 & 61.2$\vert$25.6 & 59.2$\vert$36.8 & 55.8$\vert$52.2\\
NLNL \cite{kim2019nlnl} & ICCV19 & 68.5$\vert$41.6 & 30.3$\vert$17.0 & 68.3$\vert$62.0 & \textbf{49.8}$\vert$\underline{78.1} & 57.9$\vert$\textbf{86.6} & 53.2$\vert$71.5 & 60.4$\vert$30.7 & 58.6$\vert$38.9 & 55.9$\vert$53.3\\
SCE \cite{wang2019symmetric} & ICCV19 & 64.4$\vert$37.5 & 39.0$\vert$24.6 & 65.3$\vert$60.6 & 47.5$\vert$77.7 & 44.9$\vert$76.6 & 52.8$\vert$67.5 & 61.5$\vert$26.2 & 54.7$\vert$38.5 & 53.8$\vert$51.1\\
APL \cite{ma2020normalized} & ICML20 & 69.1$\vert$38.0 & 33.4$\vert$20.0 & 67.8$\vert$64.0 & 47.2$\vert$\textbf{78.6} & 50.6$\vert$82.2 & 55.8$\vert$72.6 & 57.3$\vert$22.6 & 55.7$\vert$30.5 & 54.6$\vert$51.1\\
PolyLoss \cite{leng2022polyloss} & ICLR22 & 67.6$\vert$40.3 & 39.3$\vert$25.3 & 67.4$\vert$61.8 & 46.0$\vert$73.8 & 51.0$\vert$80.4 &  50.8$\vert$66.8 & \textbf{62.0}$\vert$27.0 & 54.4$\vert$34.9 & 54.8$\vert$51.3\\
ANL \cite{ye2023active} & NeurIPS23 & 67.4$\vert$\textbf{43.7} & 35.7$\vert$20.5 & 68.6$\vert$65.5 & 46.7$\vert$76.5 & 53.8$\vert$83.4 & 52.5$\vert$68.5 & 59.1$\vert$25.4 & \textbf{60.2}$\vert$38.4 & 55.5$\vert$52.7\\
AUL \cite{zhou2023asymmetric} & TPAMI23 & 66.5$\vert$39.4 & 24.6$\vert$20.6 & 67.6$\vert$66.2 & 42.1$\vert$\underline{78.1} & 54.9$\vert$83.4 & 51.5$\vert$\underline{73.2} & 52.6$\vert$20.6 & 58.2$\vert$35.5 & 52.3$\vert$52.1\\
\midrule
Source model only & - & 67.9$\vert$39.0 & 40.0$\vert$27.2 & 66.3$\vert$61.1 & 44.2$\vert$69.6 & 50.8$\vert$76.4 & 48.8$\vert$65.8 & 61.7$\vert$26.2 & 53.9$\vert$36.1 & 54.2$\vert$50.2\\
\textbf{FAL (ours)} & - & 64.8$\vert$41.0 & 41.6$\vert$35.2 & \underline{71.1}$\vert$\underline{71.3} & \underline{49.3}$\vert$57.3 & 70.1$\vert$83.4 & \underline{57.7}$\vert$66.6 & 53.0$\vert$\underline{48.4} & 50.3$\vert$52.4 & \textbf{57.2}$\vert$\textbf{57.0}\\
\midrule
Target-supervised & - & 90.0$\vert$77.1 & 56.9$\vert$89.4 & 83.0$\vert$86.2 & 84.5$\vert$76.9 & 80.3$\vert$90.4 & 74.1$\vert$88.6 & 55.5$\vert$83.3 & 77.2$\vert$83.1 & 75.2$\vert$84.4\\
\bottomrule
\end{tabular}
}
\label{tab:emotion}
\end{table*}

\subsection{Results of Visual Emotion Recognition}
Table \ref{tab:sentiment} shows the comparison results on the 2-class sentiment datasets. We report the classification accuracies for 12 sub-tasks and the average performance among them. From the table, we can see that: (1) Existing SFDA methods do not perform well on this dataset, with most methods even performing worse than using the source model alone. For instance, AaD \cite{yang2022attracting}'s performance drops by 7.4\% compared to the source model. This is mainly due to the significant distribution differences between visual emotion data and traditional image classification data, which causes these advanced SFDA methods to fail in the SFDA-VER task. (2) Compared to SFDA methods, the loss functions show superior performance, improving the source model's accuracy to varying degrees. As most of these loss functions have been proven to be noise-tolerant, this indicates a certain similarity between SFDA-VER and the task of learning with noisy labels. (3) Our FAL achieves significant improvements in classification accuracy, exhibiting a 3.5\% increase over the source model, and performs best or second-best in 7 out of 12 sub-tasks. This demonstrates the effectiveness of FAL in SFDA-VER.

Table \ref{tab:emotion} presents the experimental results on the 8-class emotion datasets. We provide the accuracy for each of the 8 emotion categories. To save space in the paper, we combine the domain adaptation results for EmoSet$\rightarrow$FI and FI$\rightarrow$EmoSet, which are shown on the left and right sides of the ``$|$'' symbol, respectively. From the experimental results, it can be observed that: (1) The two datasets are more challenging than the 2-class sentiment datasets, as evidenced by the significant decrease in accuracy. This is mainly because EmoSet and FI have more categories and more severe fuzzy problems, which increases the difficulty of SFDA-VER. (2) The overall performance of SFDA methods is less stable compared to loss functions. Notably, SHOT outperforms other methods in this comparison. However, some of the most advanced methods, such as C-SFDA \cite{karim2023c} and AUL \cite{zhou2023asymmetric}, do not perform well. This suggests that more complex methods may not consistently deliver optimal results, advocating for the potential superiority of simpler, more intuitive approaches. (3) Our FAL obtains  substantial performance enhancements of 3\% and 6.8\% across the two sub-tasks, surpassing all other methods. This indicates that even on more complex VER datasets, FAL can still maintain its effectiveness.

\begin{table*}[t]
\centering
\caption{Classification accuracies (\%) on Office-Home dataset (ResNet-50). We use \textbf{bold} to indicate the best results and \underline{underline} to highlight the second-best results.}
\resizebox{1.0\linewidth}{!}{
\begin{tabular}{llcccccccccccc>{\columncolor{gray!20}}c}
\toprule
Method (Source$\rightarrow$Target) & Venue & Ar$\rightarrow$Cl & Ar$\rightarrow$Pr & Ar$\rightarrow$Rw & Cl$\rightarrow$Ar & Cl$\rightarrow$Pr & Cl$\rightarrow$Rw & Pr$\rightarrow$Ar & Pr$\rightarrow$Cl & Pr$\rightarrow$Rw & Rw$\rightarrow$Ar & Rw$\rightarrow$Cl & Rw$\rightarrow$Pr & Avg. \\
\midrule
\textit{SFDA methods}\\
SHOT \cite{liang2020we} & ICML20 & 57.1 & 78.1 & 81.5 & 68.0 & 78.2 & 78.1 & 67.4 & 54.9 & 82.2 & 73.3 & 58.8 & 84.3 & 71.8 \\
G-SFDA \cite{yang2021generalized} & ICCV21 & 57.9 & 78.6 & 81.0 & 66.7 & 77.2 & 77.2 & 65.6 & 56.0 & 82.2 & 72.0 & 57.8 & 83.4 & 71.3 \\
NRC \cite{yang2021exploiting} & NeurIPS21 & 57.7 & \underline{80.3} & 82.0 & 68.1 & 79.8 & 78.6 & 65.3 & 56.4 & 83.0 & 74.0 & 58.6 & 85.6 & 72.2 \\
CPGA \cite{qiu2021source} & IJCAI21 & 59.3 & 78.1 & 79.8 & 65.4 & 75.5 & 76.4 & 65.7 & 58.0 & 81.0 & 72.0 & \textbf{64.4} & 83.3 & 71.6 \\
A$^2$Net \cite{xia2021adaptive} & ICCV21 & 58.4 & 79.0 & 82.4 & 67.5 & 79.3 & 78.9 & 68.0 & 56.2 & 82.9 & \underline{74.1} & 60.5 & 85.0 & 72.8 \\
SHOT++ \cite{liang2021source} & TPAMI21 & 57.9 & 79.7 & 82.5 & 68.5 & 79.6 & 79.3 & 68.5 & 57.0 & 83.0 & 73.7 & 60.7 & 84.9 & 73.0\\
U-SFAN \cite{roy2022uncertainty} & ECCV22 & 57.8 & 77.8 & 81.6 & 67.9 & 77.3 & 79.2 & 67.2 & 54.7 & 81.2 & 73.3 & 60.3 & 83.9 & 71.9 \\
CoWA-JMDS \cite{lee2022confidence} & ICML22 & 56.9 & 78.4 & 81.0 & 69.1 & 80.0 & \underline{79.9} & 67.7 & 57.2 & 82.4 & 72.8 & 60.5 & 84.5 & 72.5 \\
AaD \cite{yang2022attracting} & NeurIPS22 & 59.3 & 79.3 & 82.1 & 68.9 & 79.8 & 79.5 & 67.2 & 57.4 & 83.1 & 72.1 & 58.5 & 85.4 & 72.7 \\
DaC \cite{zhang2022divide} & NeurIPS22 & 59.1 & 79.5 & 81.2 & 69.3 & 78.9 & 79.2 & 67.4 & 56.4 & 82.4 & 74.0 & \underline{61.4} & 84.4 & 72.8 \\
ELR \cite{yi2023source} & ICLR23 & 58.4 & 78.7 & 81.5 & 69.2 & 79.5 & 79.3 & 66.3 & 58.0 & 82.6 & 73.4 & 59.8 & 85.1 & 72.6 \\
C-SFDA \cite{karim2023c} & CVPR23 & 60.3 & 80.2 & \underline{82.9} & 69.3 & \underline{80.1} & 78.8 & 67.3 & \underline{58.1} & \textbf{83.4} & 73.6 & 61.3 & \underline{86.3} & \textbf{73.5} \\
SHOT+DPC \cite{xia2024discriminative} & CVPR24 & 59.2 & 79.8 & 82.6 & 68.9 & 79.7 & 79.5 & \underline{68.6} & 56.5 & 82.9 & 73.9 & 61.2 & 85.4 & 73.2\\
Improved SFDA \cite{mitsuzumi2024understanding} & CVPR24 & \textbf{60.7} & 78.9 & 82.0 & \underline{69.9} & 79.5 & 79.7 & 67.1 & \textbf{58.8} & 82.3 & \textbf{74.2} & 61.3 & \textbf{86.4} & \underline{73.4}\\
\midrule
\textit{loss functions}\\
Focal loss \cite{lin2017focal} & ICCV17 & 45.9 & 68.5 & 75.4 & 55.7 & 64.7 & 68.9 & 55.8 & 44.4 & 75.5 & 64.9 & 46.2 & 78.6 & 62.0\\
GCE \cite{zhang2018generalized} & NeurIPS18 & 50.1 & 74.1 & 77.1 & 62.6 & 71.2 & 72.6 & 59.2 & 46.1 & 78.4 & 67.6 & 49.4 & 80.8 & 65.8\\
NLNL \cite{kim2019nlnl} & ICCV19 & 44.3 & 64.8 & 73.9 & 56.9 & 64.5 & 66.8 & 54.7 & 42.9 & 74.5 & 66.4 & 48.6 & 77.2 & 61.3\\
SCE \cite{wang2019symmetric} & ICCV19 & 45.5 & 66.5 & 72.5 & 52.3 & 63.1 & 66.3 & 53.8 & 42.6 & 71.7 & 63.4 & 45.6 & 77.5 & 60.1\\
APL \cite{ma2020normalized} & ICML20 & 49.0 & 74.0 & 78.2 & 61.7 & 70.2 & 72.2 & 59.1 & 45.6 & 77.9 & 67.2 & 46.0 & 81.6 & 65.2\\
PolyLoss \cite{leng2022polyloss} & ICLR22 & 47.0 & 69.8 & 75.9 & 57.3 & 66.1 & 69.8 & 56.9 & 44.9 & 76.0 & 65.5 & 46.6 & 79.5 & 62.9\\
ANL \cite{ye2023active} & NeurIPS23 & 47.4 & 69.1 & 75.4 & 58.7 & 67.3 & 69.2 & 57.5 & 45.2 & 76.3 & 67.4 & 50.8 & 78.7 & 63.6\\
AUL \cite{zhou2023asymmetric} & TPAMI23 & 47.1 & 72.5 & 76.3 & 58.2 & 69.2 & 68.3 & 56.1 & 42.9 & 75.4 & 64.9 & 45.6 & 80.5 & 63.1\\
\midrule
Source model only & - & 44.6 & 67.3 & 74.8 & 52.7 & 62.7 & 64.8 & 53.0 & 40.6 & 73.2 & 65.3 & 45.4 & 78.0 & 60.2\\
\textbf{FAL (ours)} & - & 56.5 & 79.5 & 81.6 & 67.9 & 79.1 & 78.8 & 67.7 & 53.9 & 82.0 & 72.6 & 59.0 & 84.8 & 72.0\\
\textbf{FAL++ (ours)} & - & 58.3 & \textbf{81.1} & \textbf{83.1} & \textbf{70.0} & \textbf{81.3} & \textbf{80.9} & \textbf{69.6} & 55.1 & \underline{83.3} & 73.3 & 59.6 & 85.7 & \underline{73.4}\\
\midrule
Target-supervised & - & 77.9 & 91.4 & 84.4 & 74.5 & 91.4 & 84.4 & 74.5 & 77.9 & 84.4 & 74.5 & 77.9 & 91.4 & 82.0\\
\bottomrule
\end{tabular}
}
\label{tab:officehome}
\end{table*}

\subsection{Results of Object Recognition}
In addition to VER datasets, we also evaluate the performance of FAL and other loss functions on the highly challenging Office-Home dataset. Similar to the 2-class sentiment datasets, we report the results on 12 sub-tasks. Considering that many SFDA methods combine some auxiliary techniques to achieve better performance, we combine FAL with the labeling transfer strategy proposed in SHOT++ \cite{liang2021source} to obtain an enhanced version, i.e., FAL++. From Table \ref{tab:officehome}, we can see that: (1) The loss functions perform much worse than advanced SFDA methods on this dataset, which contrasts sharply with its performance on VER datasets. This further highlights the significant differences between SFDA-VER and the conventional SFDA task. (2) Our FAL significantly outperforms all other loss functions, achieving an accuracy improvement of 11.8\% compared to the source model. Additionally, FAL++ achieves performance that is comparable to state-of-the-art SFDA methods. This indicates that, although FAL is designed for VER, it is also applicable to standard SFDA tasks, demonstrating its stronger generalization ability compared to existing SFDA methods and loss functions.

\subsection{Analysis and Discussions}
\textbf{Ablation study}. We investigate the advantages of each component of the proposed loss function in Table \ref{tab:ablation}. $\ell_{fal-1}$ and $\ell_{fal-2}$ denote the first and second terms of FAL in Eq. (\ref{eq:faloss3}), and weights refer to the $w$ in Eq. (\ref{eq:faloss2}). The results clearly show that $\ell_{fal-1}$ has a significant effect, and the performance is further improved by adding $\ell_{fal-2}$. After incorporating weights, there is a slight improvement in accuracy. It is worth noting that using $\ell_{fal-2}$ alone can lead to performance degradation as $\ell_{fal-2}$ primarily provides robustness to noise but lacks high-level semantic guidance for emotion categories. This is consistent with the observations in works such as SHOT \cite{liang2020we}, AaD \cite{yang2022attracting}, SCE \cite{wang2019symmetric}, and AUL \cite{zhou2023asymmetric}, which typically require combining two loss terms to achieve better learning results.

\begin{table}[htbp]
\centering
\caption{Average accuracies (\%) on the 8-class emotion datasets.}
\begin{tabular}{lcc}
\toprule
{Methods} & {EmoSet$\rightarrow$FI} & {FI$\rightarrow$EmoSet}\\
\midrule
Source model only & 54.2 & 50.2 \\
$\ell_{fal-1}$ & 56.5 & 55.5 \\
$\ell_{fal-2}$ & 51.0 & 45.9 \\
$\ell_{fal-1}$ $+$ $\ell_{fal-2}$ & 57.1 & 56.7 \\
$\ell_{fal-1}$ $+$ $\ell_{fal-2}$ $+$ weights & 57.2 & 57.0 \\
\bottomrule
\end{tabular}
\label{tab:ablation}
\end{table}

\begin{figure}[t]
\centering
\footnotesize
    \begin{tabular}{cc}
    \hspace{-10pt}
    \subfloat[Source model (EmoSet$\rightarrow$FI)]{
    \includegraphics[width=0.22\textwidth]{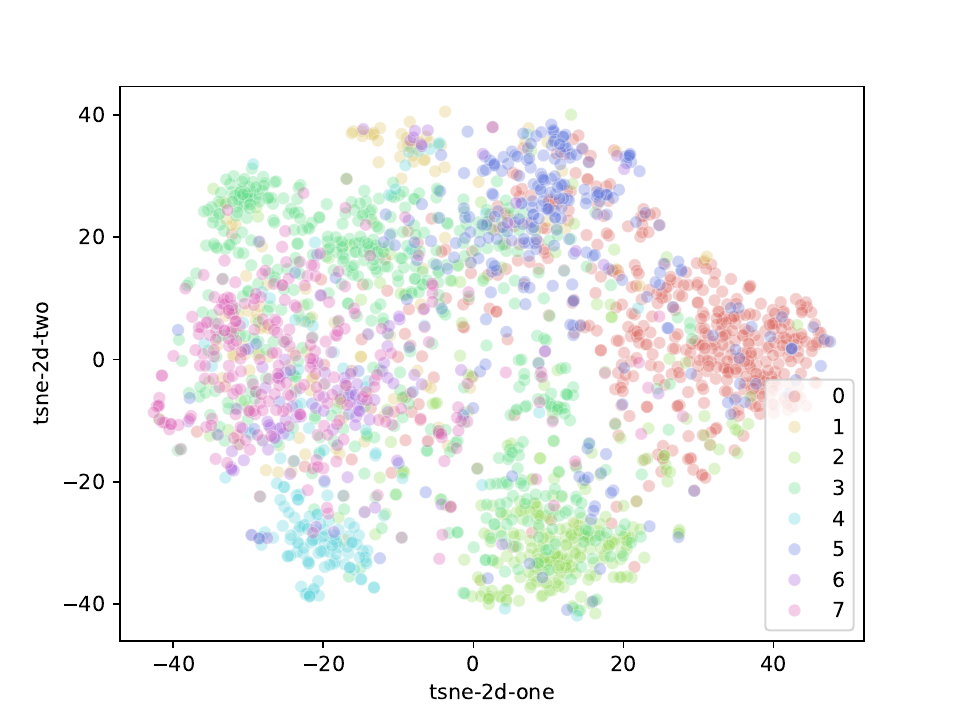}
    \label{fig:subfig1}
    } &
    \hspace{-10pt}
    \subfloat[Source model (FI$\rightarrow$EmoSet)]{
    \includegraphics[width=0.22\textwidth]{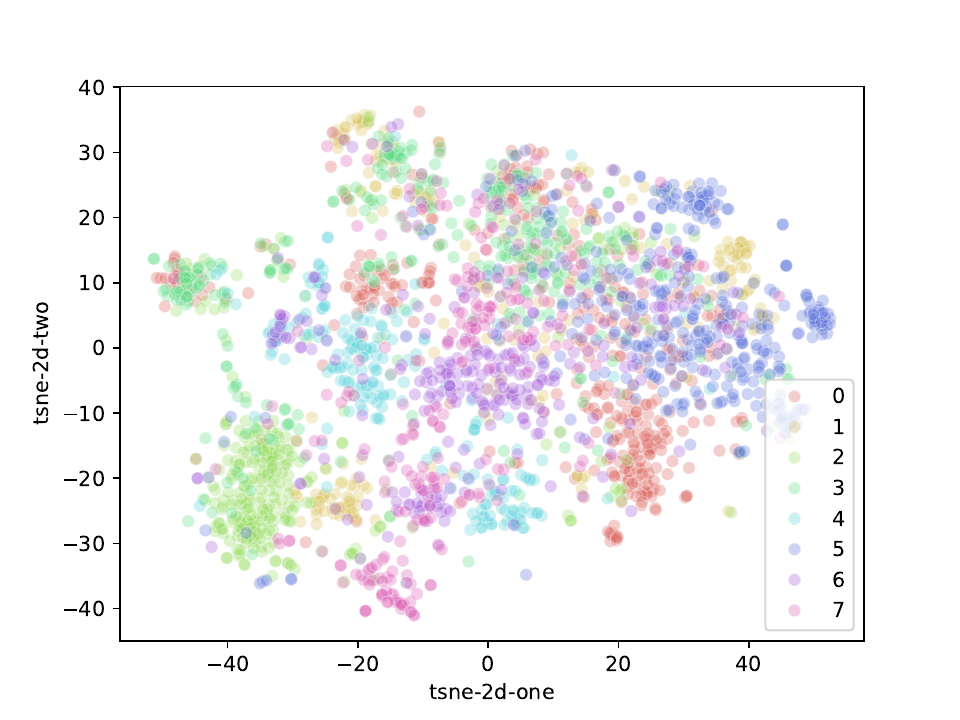}
    \label{fig:subfig2}
    } \\
    \hspace{-10pt}
    \subfloat[Our FAL (EmoSet$\rightarrow$FI)]{
    \includegraphics[width=0.22\textwidth]{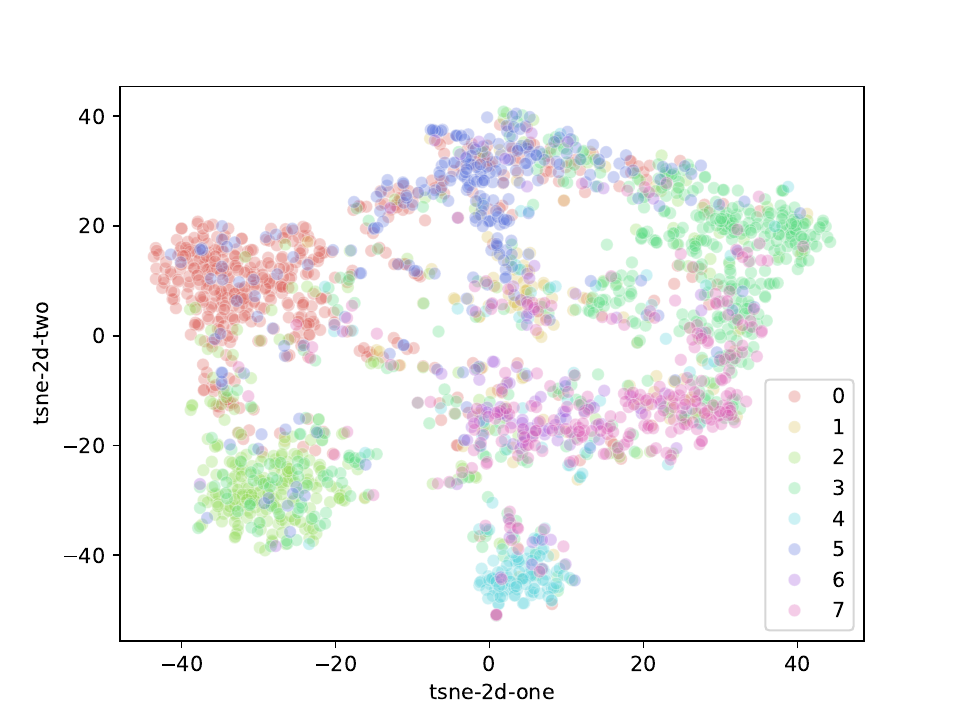}
    \label{fig:subfig3}
    } &
    \hspace{-10pt}
    \subfloat[Our FAL (FI$\rightarrow$EmoSet)]{
    \includegraphics[width=0.22\textwidth]{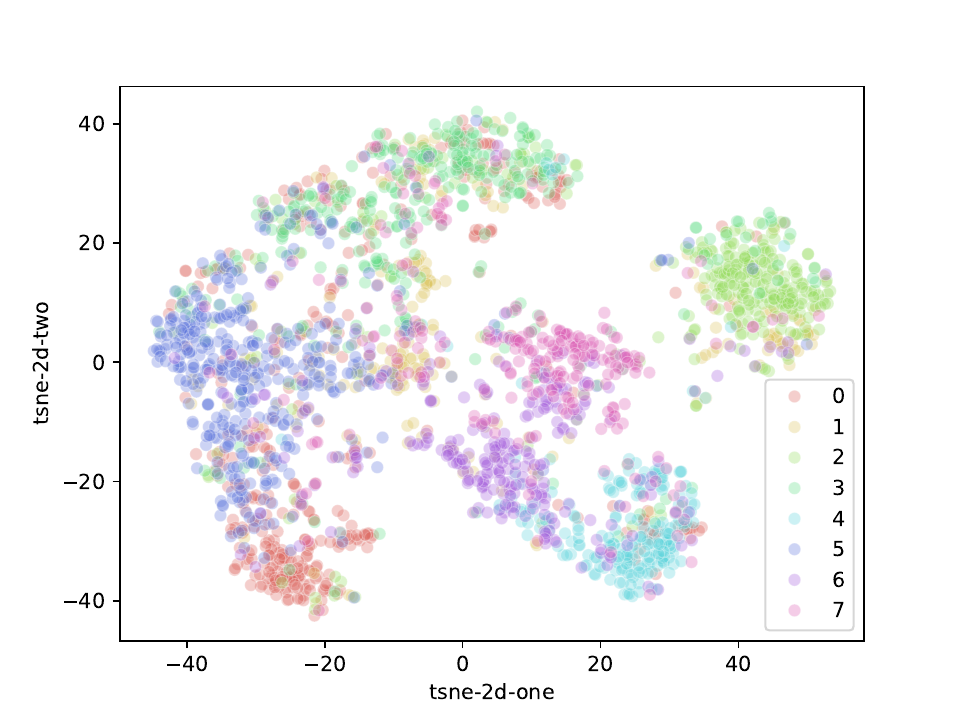}
    \label{fig:subfig4}
    }
    \end{tabular}
\caption{Visualization of feature representations learned by the source model (top) and our FAL (bottom) on two tasks of the 8-class emotion datasets.}
\label{fig:tsne}
\end{figure}

\textbf{Feature visualization}. We further compare the visualization results of the feature representations learned by the source model and our FAL. We first extract the high-dimensional features output by the second-to-last layer, i.e., the bottleneck layer, and then utilize t-SNE \cite{van2008visualizing} to project these features into 2D embeddings. Considering that the dataset size of EmoSet and FI is too large, displaying all of them would be too messy. To visualize more clearly, we uniformly sample 1/10 and 1/50 samples on the EmoSet and FI datasets, respectively. The visualization results of the feature representations are shown in Fig. \ref{fig:tsne}. We can see that the feature representations extracted by the source model on the target domain are not discriminative enough, with different categories mixed together. In contrast, the feature representations learned by our FAL have clearer boundaries and more concentrated clusters, indicating a significant improvement in feature quality compared to the source model.

\section{Conclusion}
\label{sec:conclusion}
In this work, we delve into the SFDA-VER task from a fuzzy perspective and identify two prominent issues: fuzzy emotion labels and fuzzy pseudo-labels. To tackle these issues, we propose a novel objective function called fuzzy-aware loss (FAL). FAL integrates the standard cross entropy loss with a fuzzy-aware loss term, aiming to calibrate the loss of non-predicted classes to accommodate fuzzy labels. Furthermore, we provide theoretical proof of FAL's robustness to noise and discuss the relationship between FAL and two typical loss functions. Extensive experimental results on three benchmark datasets show that FAL achieves state-of-the-art accuracy. We hope that this work can lay the foundation for the future development of SFDA-VER and promote the practical application of VER technology.

\bibliographystyle{IEEEtran}
\bibliography{refs}

% \newpage

\begin{IEEEbiographynophoto}{Ying Zheng} is currently a Research Fellow at The Hong Kong Polytechnic University, Hong Kong, China. Before that, he received the Ph.D. degree from Harbin Institute of Technology, Harbin, China. His research interests include machine learning, computer vision, and embodied AI.
\end{IEEEbiographynophoto}

\vspace{-22pt}

\begin{IEEEbiographynophoto}{Yiyi Zhang}
received her B.Eng. degree from Zhejiang University and M.Sc. degree from the Institut Polytechnique de Paris - Télécom Paris. She is currently working towards the Ph.D. degree with The Chinese University of Hong Kong. Her research interests include Computer Vision and Medical AI.
\end{IEEEbiographynophoto}

\vspace{-22pt}

\begin{IEEEbiographynophoto}{Yi Wang}
(Member, IEEE) received BEng degree in electronic information engineering and MEng degree in information and signal processing from the School of Electronics and Information, Northwestern Polytechnical University, Xi’an, China, in 2013 and 2016, respectively. He earned PhD in the School of Electrical and Electronic Engineering from Nanyang Technological University, Singapore, in 2021. He is currently a Research Assistant Professor at the Department of Electrical and Electronic Engineering, The Hong Kong Polytechnic University, Hong Kong. His research interest includes Image/Video Processing, Computer Vision, Intelligent Transport Systems, and Digital Forensics.
\end{IEEEbiographynophoto}

\vspace{-22pt}

\begin{IEEEbiographynophoto}{Lap-Pui Chau}
 (Fellow, IEEE) received a Ph.D. degree from The Hong Kong Polytechnic University in 1997. He was with the School of Electrical and Electronic Engineering, Nanyang Technological University from 1997 to 2022. He is currently a Professor in the Department of Electrical and Electronic Engineering, The Hong Kong Polytechnic University. His current research interests include large language model, perception for autonomous driving, egocentric computer vision, and 3D computer vision. He is an IEEE Fellow. He was the chair of Technical Committee on Circuits \& Systems for Communications of IEEE Circuits and Systems Society from 2010 to 2012. He was general chairs and program chairs for some international conferences. Besides, he served as associate editors for several IEEE journals and Distinguished Lecturer for IEEE BTS. 
\end{IEEEbiographynophoto}

\vfill

\end{document}